\documentclass{article}
\usepackage{ijcai25}

\usepackage{times}
\usepackage{soul}
\usepackage{url}
\usepackage[hidelinks]{hyperref}
\usepackage[utf8]{inputenc}
\usepackage[small]{caption}
\usepackage{graphicx}
\usepackage{amsmath}
\usepackage{amsthm}
\usepackage{booktabs}
\usepackage{algorithm}
\usepackage{algorithmic}
\usepackage[switch]{lineno}
\usepackage{amssymb}

\usepackage{xcolor}

\title{ Safety of Embodied Navigation: A Survey}

\author{
Zixia Wang$^1$
\and
Jia Hu$^1$
\and
Ronghui Mu$^{1}$\thanks{Corresponding Author} \\
\affiliations
$^1$University of Exeter\\
\emails
\{zw483,j.hu,r.mu2\}@exeter.ac.uk
}

\begin{document}

\maketitle

\begin{abstract}
As large language models (LLMs) continue to advance and gain influence, the development of embodied AI has accelerated, drawing significant attention, particularly in navigation scenarios. Embodied navigation requires an agent to perceive, interact with, and adapt to its environment while moving toward a specified target in unfamiliar settings. However, the integration of embodied navigation into critical applications raises substantial safety concerns. Given their deployment in dynamic, real-world environments, ensuring the safety of such systems is critical. This survey provides a comprehensive analysis of safety in embodied navigation from multiple perspectives, encompassing attack strategies, defense mechanisms, and evaluation methodologies. Beyond conducting a comprehensive examination of existing safety challenges, mitigation technologies, and various datasets and metrics that assess effectiveness and robustness, we explore unresolved issues and future research directions in embodied navigation safety. These include potential attack methods, mitigation strategies, more reliable evaluation techniques, and the implementation of verification frameworks. By addressing these critical gaps, this survey aims to provide valuable insights that can guide future research toward the development of safer and more reliable embodied navigation systems. Furthermore, the findings of this study have broader implications for enhancing societal safety and increasing industrial efficiency.
\end{abstract}
\section{Introduction}
\begin{figure}[ht] 
    \centering 
\includegraphics[width=0.48\textwidth]{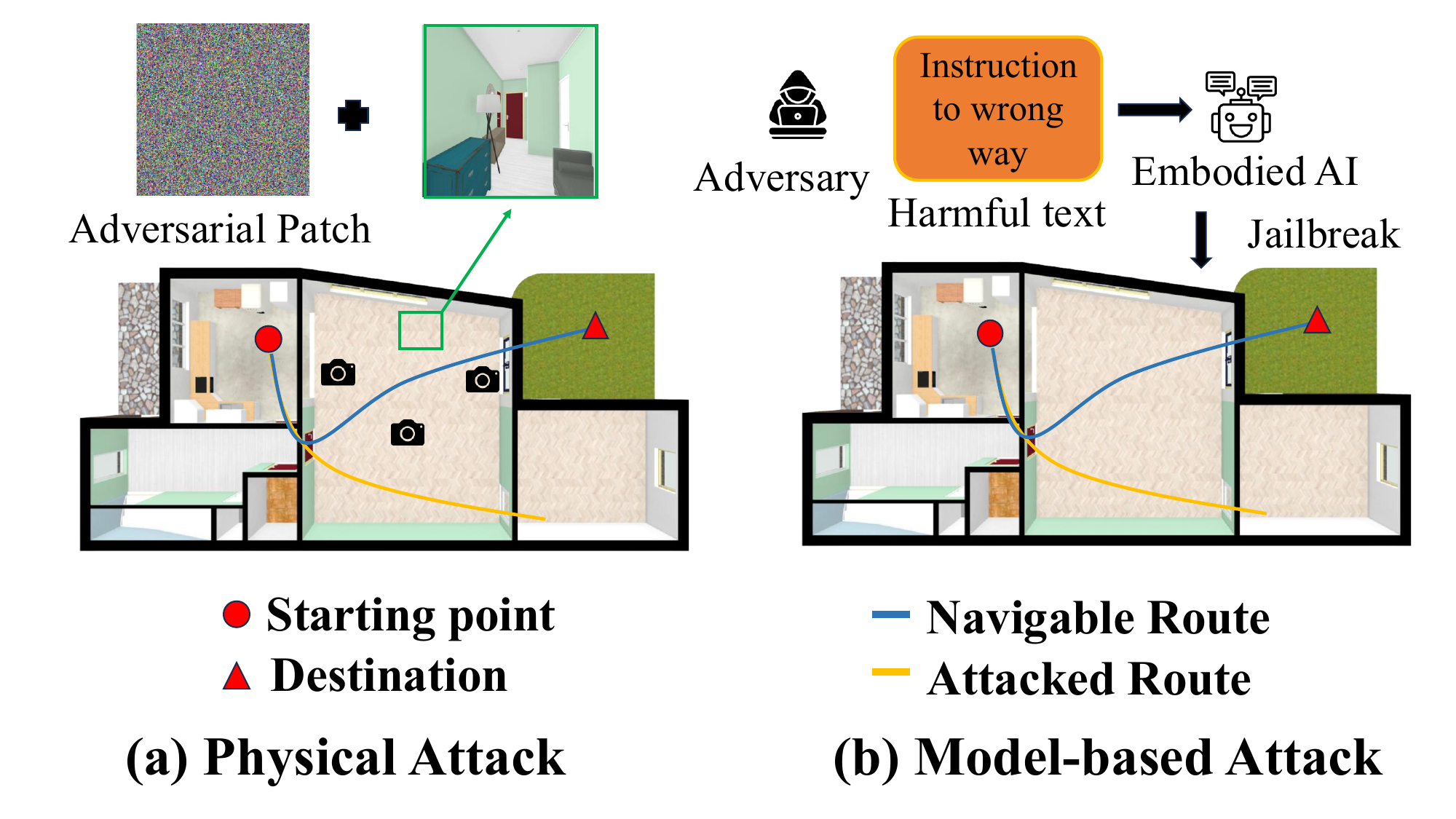}
    \caption{
Examples of two types of attacks. The red circle marks the starting point, the red triangle indicates the destination, the green line represents the navigable route, and the yellow line shows the attacked route. In the physical attack example, an adversarial patch disrupts navigation, while in the model-based attack example, a jailbreak injects harmful instructions, leading to incorrect actions.} 
    \label{fig:example} 
\end{figure} 
In recent years, Large Language Models (LLMs) have garnered significant attention for their remarkable capabilities in perception, interaction, and reasoning. These advancements have contributed to the rise of Embodied Artificial Intelligence (Embodied AI), which serves as a bridge between the virtual and physical worlds. 
A key aspect of Embodied AI is embodied navigation, which enables an AI agent to perceive and interact with its environment while moving toward a target or specified location in unfamiliar settings. This requires a combination of intelligent capabilities, including visual perception, mapping, planning, exploration, and reasoning.  For example, consider an AI agent instructed to \textit{``Retrieve a bottle of water from the kitchen fridge.''} The agent must navigate to the kitchen, identify the fridge, pick up the correct item, and return to the designated location.  Embodied navigation plays a crucial role in various real-world applications, including safety-critical scenarios such as robotic navigation \cite{wang2024divscenebenchmarkinglvlmsobject} and autonomous driving \cite{li2023benchmarkingassessingvisualnaturalness}. Therefore, ensuring the safety and efficiency of embodied navigation is crucial.

However, the security of Embodied AI remains a significant concern, as it relies on deep neural networks (DNNs), which are susceptible to adversarial attacks \cite{liu2020spatiotemporalattacksembodiedagents}. These vulnerabilities pose serious risks to the safety and reliability of embodied navigation systems. One type of attack alters the physical environment to mislead navigation perception. For example, adversarial patches or perturbations placed on objects or surfaces can mislead the input of the model, causing it to misunderstand its surroundings \cite{chen2024towards}. Another form of attack directly targets the AI model by injecting maliciously crafted inputs that manipulate its decision-making process \cite{liu2024exploringrobustnessdecisionleveladversarial}. For example, carefully designed adversarial prompts can trick a large model into producing incorrect or harmful outputs. As a result, the agent may misidentify barriers, take incorrect paths, or even crash into objects (as shown in Figure \ref{fig:example}, both physical patch attacks and model-based jailbreak attacks cause embodied navigation to deviate from its original correct path).  While some existing studies focus on defense methods \cite{wu2024embodiedactivedefenseleveraging} or the construction of safety-related benchmarks \cite{yin2024safeagentbenchbenchmarksafetask}, a comprehensive survey on the safety of embodied navigation is still lacking, which limits a holistic understanding of the field’s development.

\begin{figure}[htb] 
    \centering 
    \includegraphics[width=0.48\textwidth]{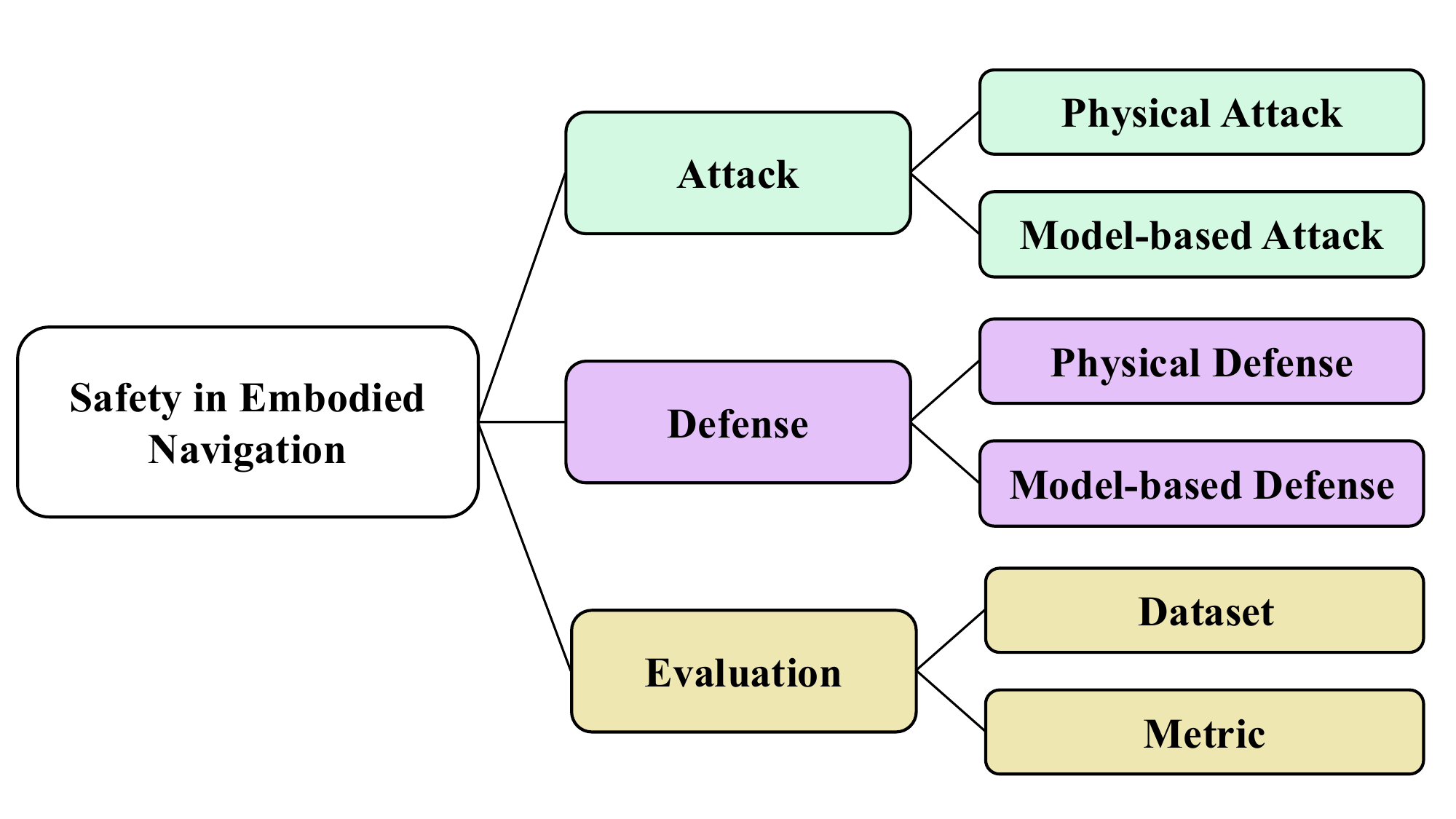}
    \caption{
    Taxonomy: Safety of Embodied Navigation
} 
    \label{fig:tax} 
\end{figure}

In this paper, we aim to explore three key research questions: (1) \textit{What risks do embodied navigation agents face?} (2) \textit{What methods can be employed to mitigate these risks and enhance system reliability?} (3) \textit{What metrics can be used to evaluate the safety of embodied navigation agents?}  To address these questions, we present a comprehensive survey that summarizes recent advancements in three critical areas: \textbf{attacks, defenses, and evaluation} (as illustrated in Figure~\ref{fig:tax}). First, after comprehensive investigation the safety risks, we categorize potential threats into two primary types: \textbf{physical attacks}, which are caused by environmental factors such as adversarial patches or lighting conditions, and \textbf{model-based attacks}, which exploit vulnerabilities in the navigation model itself, particularly in large-scale models. These attack types are further classified based on their nature, the attackers involved, and the methodologies employed. Next, we systematically examine existing defense mechanisms for embodied navigation, aligning them with the corresponding attack types to provide a structured understanding of adversarial threats and mitigation strategies. Additionally, we emphasize the critical role of well-structured datasets and robust evaluation metrics in ensuring the safety and reliability of embodied navigation systems. Given that evaluation is a fundamental aspect of safety assessment, we discuss the necessity of standardized benchmarks and comprehensive testing methodologies in advancing the security of embodied AI.

Building on our comprehensive review of existing research, we present some future directions in embodied AI safety. These include advancing attack strategies, particularly in multimodal AI settings; developing more effective and adaptive defense mechanisms for real-time navigation; and establishing standardized evaluation frameworks to enhance fairness and interpretability across diverse tasks. Moreover, we highlight the importance of verification techniques in quantifying robustness thresholds and defining theoretical performance bounds. We believe these efforts will contribute to the development of safer and more reliable embodied AI systems.
Notably, ~\cite{zhang2024visionandlanguagenavigationtodaytomorrow,liu2025screensscenessurveyembodied} are works related to us, focusing primarily on embodied navigation (excluding safety aspects) and safety within embodied AI in one specific area (healthcare), respectively. In contrast, the central focus of our work is on safety issues within embodied navigation across a general domain. 
Our main contributions are as follows:

\begin{itemize}  
    \item We provide a systematic review of attack and defense methods related to the safety of embodied navigation. To the best of our knowledge, this is the first comprehensive study on the safety of embodied navigation.
    \item We compare and analyze recent evaluation datasets and metrics used for assessing embodied navigation safety.  
    \item We present potential future research directions in embodied navigation safety to inform and inspire further advancements toward the development of more robust and reliable embodied navigation systems.
\end{itemize}

\section{Preliminaries}

In this section, we introduce the background knowledge in embodied navigation and safety to better illustrate the scope of this survey. 

\subsection{Overview of Embodied Navigation}
Embodied AI refers to intelligent systems that integrate perception, decision-making, and action within a physical or simulated environment, enabling them to interact with and adapt to their surroundings autonomously. A crucial application of embodied AI is embodied navigation, which enables an agent to perceive its surroundings, plan a path, and move toward a target while adapting to dynamic and unfamiliar settings.
In an embodied navigation task, an agent is placed in a visual environment \(E\) and given an instruction \(I\) to find a route \(R\) from a starting point \(S\) to a destination \(D\). The route \(R\) is a sequence of viewpoints that the agent will follow. At each time step \(t\), the agent observes multiple views \(\{V_{t,i}\}\); some of these views indicate possible directions of movement. Using the instruction \(I\), its previous observations, and previous commands \(\{c_0, c_1, \ldots, c_{t-1}\}\), the agent selects the next command \(c_t\). The process ends when the agent chooses the stop command, denoted as \(c_{\text{stop}}\). Our survey builds upon and extends this foundational task.

Embodied navigation focuses on enabling agents to move and interact in physical environments using visual and sensory input. Key tasks include object goal navigation (e.g., locating specific objects like ``kitchen'' in unknown settings~\cite{chen2024towards,Ying_2023,yang2024hijackingvisionandlanguagenavigationagents}), image goal navigation (reaching a target depicted in an image like ``fridge'' 
\cite{mezghani2022memoryaugmentedreinforcementlearningimagegoal}) , visual language navigation (VLN) (following natural language instructions to traverse a route~\cite{jiao2024trustembodiedagentsexploring,lu2024poexpolicyexecutableembodied}) and interactive navigation (answering questions about the environment \cite{liu2024exploringrobustnessdecisionleveladversarial}).
\begin{table*}[htbp]
\centering
\resizebox{\textwidth}{!}{%
\begin{tabular}{@{}cccccc@{}}
\toprule
\textbf{Type} & \textbf{Paper} & \textbf{Attacker} & \textbf{Attack Type} & \textbf{Technique} & \textbf{Brief Description} \\
\midrule
 &  \cite{Ying_2023} & User & White-box & 2D-patch & Constant image-agnostic perturbation  applied to each input frame \\
 & \cite{wei2023distributionalmodelinglocationawareadversarial} & User & Black-box & 2D-patch & Optimizing adversarial patch placement for optimal positioning
 \\
 & \cite{suryanto2023activehighlytransferable3d} & User & Black-box & 3D-patch & General-purpose adversarial patches for vehicle applications \\
 & 
\cite{yang2024hijackingvisionandlanguagenavigationagents} & User & White-box & 3D-patch & Optimized the appearance of 3D adversarial objects \\
 &
 \cite{huang2024transferabletargeted3dadversarial} & User & Black-box & 3D-patch & Transferable targeted 3D adversarial examples \\
 Physical Attack& \cite{hu2023physically} & User & White-box & Multi-view-patch & Adversarial texture for clothes  \\
 & \cite{chen2024towards} & User & White-box & Multi-view-patch & Multiview optimization strategy based on object-aware sampling \\
 
 & \cite{hu2023adversariallaserspotrobust} & User & Black-box &  Adversarial Light & Optimizes the physical parameters of laser spots to perform physical attacks \\
 & \cite{zhang2024understandingimpactselectromagneticsignal} & Third Party  & White-box &  Adversarial Light & Electromagnetic Signal Injection Attacks \\
 & \cite{sun2024embodiedlaserattackleveragingscene} & User, Third Party & Black-box &Adversarial Light & Dynamically tailored non-contact laser attack \\
 
\midrule

& \cite{sun2024embodiedlaserattackleveragingscene} & User, Third Party & Black-box &Reinforcement Learning & Dynamically tailored non-contact laser attack \\
 & \cite{zhang2024navigationattackerswishbuilding}  & User & Black-box & Federated Learning & Backdoor attack on FL-based embodied agents\\
 & \cite{jiao2024trustembodiedagentsexploring} 
 & User & White-box & LLM-Backdoor attack & Backdoor attacks against embodied LLM-based decision-making systems \\
 Model-based Attack & \cite{wang2025trojanrobotphysicalworldbackdoorattacks}& User, Third Party & Black-box & LLM-Backdoor attack &  Backdoor attacks on VLM-based robotic manipulation \\
& \cite{zhang2024badrobotmanipulatingembodiedllms}   & Third Party & Black-box & LLM-Jailbreak attack & Manipulation, alignment, and knowledge for attacks \\
 & \cite{lu2024poexpolicyexecutableembodied}
 & User & Black-box & LLM-Jailbreak attack &  Adversarial and meaningful suffixes with a focus on simple words \\
 & \cite{liu2024exploringrobustnessdecisionleveladversarial} & User & White-box &  LLM-Jailbreak attack & Targeted attacks for controlled manipulation and untargeted attacks for random disruptions  \\

\bottomrule
\end{tabular}
}
\caption{Different types of representative attacks: ``Attacker'' refers to the adversary, categorized into user and third party; ``Attack type'' is classified into black-box and white-box attacks, while ``Technique'' corresponds to different attack methodologies. }
\label{tab:booktabs}
\end{table*}

\subsection{Safety in Embodied Intelligence}

Currently, research on embodied AI safety remains relatively limited, with different studies using various metrics to assess these systems. In general, safety is understood as a model’s ability to withstand perturbations, where greater robustness means stronger resistance to disruption. Specifically, adversarial perturbations pose a significant challenge to embodied navigation by distorting perception and decision-making. At time step \( t \), the perturbed observation is defined as \( V'_{t,i} = V_{t,i} + \delta_{t,i} \), where \( \delta_{t,i} \) represents an adversarial perturbation applied to the original observation \( V_{t,i} \), which can alter the agent’s perception of the environment. Humans are usually unable to detect these disturbances. Given an input \( V_{t,i} \), a model \( F \) typically produces an output \( c_t \) that aligns with human expectations. However, when \( V'_{t,i} \) is presented instead, the model's response may deviate significantly, leading to unexpected or even harmful consequences. As a result, instead of selecting the correct command \( c_t \) based on the original observations, the agent chooses \( c_t^\prime \) under the perturbed observations, where \( c_t^\prime \neq c_t \). This incorrect decision may cause the agent to deviate from its intended route, leading to navigation failure and potentially unsafe outcomes.

Embodied intelligence systems face two major security threats: physical threats and model threats. Since deep neural networks (DNNs) are highly sensitive to such disruptions, external factors such as malicious patches~\cite{Ying_2023,chen2024towards} or changes in lighting conditions~\cite{zhang2024understandingimpactselectromagneticsignal,sun2024embodiedlaserattackleveragingscene} can distort the system’s understanding of its surroundings, leading to navigation errors. On the other hand, model threats arise from vulnerabilities within large language models (LLMs). Attackers may inject malicious instructions or exploit hallucination effects~\cite{jiao2024trustembodiedagentsexploring,wang2025trojanrobotphysicalworldbackdoorattacks,zhang2024badrobotmanipulatingembodiedllms,huang2024survey,dong2024safeguarding}, potentially causing the system to generate inaccurate or even risky route plans. A detailed discussion of these threats will be presented in Chapter \ref{sec:3}.

\section{Attack}\label{sec:3}

In this section, we examine attacks on embodied navigation, broadly classified into two categories: physical attacks and model-based attacks, with some attacks combining elements of both. Table \ref{tab:booktabs} presents different types of representative attacks. However, research on the vulnerabilities of embodied systems remains limited. To address this gap, we discuss potential future attack vectors in Chapter \ref{sec:6}.

\subsection{Physical Attack}
 Since the introduction of achievable physical adversarial samples \cite{kurakin2017adversarialexamplesphysicalworld}, physical adversarial attacks widely applied in various computer vision tasks, such as facial recognition ~\cite{wei2023jailbrokendoesllmsafety} and traffic sign detection \cite{suryanto2023activehighlytransferable3d}.
 
However, research on adversarial attacks targeting embodied navigation agents remains relatively limited. To establish a foundation for understanding these threats, we first review physical patch attacks in 2D pixel spaces before exploring their extensions to 3D settings. The early exploration of physical attacks was initiated by \cite{brown2018adversarialpatch}, who proposed a method for generating universal adversarial image patches capable of attacking any scene. Later, \cite{Chen_2019} introduced adversarial perturbations on stop signs, causing Faster R-CNN to misclassify them and threatening autonomous vehicles. Furthermore, some studies ~\cite{wei2023distributionalmodelinglocationawareadversarial,li2021generativedynamicpatchattack} focused on optimizing patch placement to enhance attack effectiveness. In embodied navigation attacks, the attack is achieved by \cite{Ying_2023} through the application of agnostic perturbations to each input frame.

Due to the inherent limitations of 2D patches, researchers have redirected their attention to investigating the impact of patch-based attacks on 3D properties, such as rotation and translation \cite{zeng2019adversarialattacksimagespace}. \cite{liu2020spatiotemporalattacksembodiedagents} was an early work on attacks against embodied navigation agents, which focused on the physical attributes of objects in key scene views, such as textures and 3D shapes. Meanwhile, \cite{athalye2018synthesizingrobustadversarialexamples,wiyatno2019physicaladversarialtexturesfool} introduced the Expectation Over Transformation (EOT) method, which generated robust 3D adversarial examples by simulating real-world variations, including rotation, scaling, and blurring.


Influenced by methods such as EOT, numerous attacks have emerged. \cite{xu2020adversarialtshirtevadingperson} proposed attaching adversarial patches to t-shirts to generate robust adversarial samples that simulate deformation effects. Similarly, \cite{yang2024hijackingvisionandlanguagenavigationagents} explored the appearance optimization of 3D adversarial objects, aiming to manipulate how objects are perceived in the environment and induce the desired behavior in a pretrained VLN agent, thereby achieving an attack in embodied navigation. Additionally, some methods explored 3D adversarial camouflages by disrupting object textures within navigation scenes, particularly in autonomous driving. \cite{huang2024transferabletargeted3dadversarial} developed more effective and transferable techniques for generating targeted 3D adversarial examples. \cite{suryanto2023activehighlytransferable3d} proposed camouflage attack methods based on texture rendering, while \cite{wang2021fcalearning3dfullcoverage} implemented attacks on the full 3D surface of vehicles.

Early research on attacks against embodied navigation agents was constrained by viewpoint variations and environmental complexity, limiting their effectiveness in real-world settings. To address these challenges, recent studies have developed specialized attack strategies tailored for navigation tasks. Adversarial textures for clothing \cite{hu2023physically} were initially explored to enhance robustness against viewpoint variations. Similarly, \cite{chen2024towards} also leveraged a multi-view approach, proposing a method that attaches adversarial patches with learnable textures and opacities to objects, integrating multiple viewpoints to achieve physical attacks.

In addition to patch-based physical attacks, some adversarial attack strategies targeting navigation shifted their focus to adversarial light. Initially, this approach was applied to image classification, where a projector altered physical light conditions to deceive classifiers \cite{Huang_2022}. Later, adversarial light techniques were extended to embodied navigation. Geometric light attacks distorted entire images, leading to the misinterpretation of navigation signs by vehicles \cite{hu2023adversariallaserspotrobust}. Similarly, electromagnetic signal injection manipulated visual inputs, affecting both classification and navigation tasks \cite{zhang2024understandingimpactselectromagneticsignal}. Furthermore, laser emitters were used to attack embodied navigation agents by exploiting vulnerabilities in their perception systems \cite{sun2024embodiedlaserattackleveragingscene}.

\subsection{Model-based Attack}

In model-based attacks, unlike physical attacks that focus on changes in the environment, these attacks are directed specifically at the model itself. 

In embodied navigation tasks, reinforcement learning (RL) can be employed to train agents on how to navigate effectively. Similar to models trained in standard gaming environments, embodied navigation systems are also susceptible to certain attacks based on reinforcement learning~\cite{mu2024reward}. Rather than using time-consuming heuristic algorithms, \cite{sun2024embodiedlaserattackleveragingscene} employed reinforcement learning to optimize adversarial laser attack strategies, improving efficiency.

Federated Learning (FL) enables multiple clients, such as household navigation environments, to collaboratively train navigation models without sharing raw data with a central server, thereby preserving data privacy. In the context of embodied navigation, FL facilitates decentralized learning, allowing agents to adapt to different environments while maintaining data security. However, the decentralized nature of FL also introduces security risks. The opacity of local training processes makes the system vulnerable to adversarial manipulation~\cite{lyu2022privacyrobustnessfederatedlearning}. To explore these vulnerabilities, \cite{zhang2024navigationattackerswishbuilding} investigated how malicious clients could manipulate their local training data, allowing attackers to control the global model under specific conditions.

Large Language Models (LLMs) demonstrated immense potential for navigation in embodied artificial intelligence. With extensive common sense and advanced reasoning capabilities, these models enabled robots to better comprehend complex language commands and execute high-level tasks with greater understanding and adaptability~\cite{sharan2023llmassistenhancingclosedloopplanning}.  However, large models also faced several security issues, primarily jailbreak attacks and backdoor attacks. Jailbreak attacks exploit model vulnerabilities to bypass safety mechanisms, allowing attackers to generate restricted content using crafted prompts. Backdoor attacks embed hidden triggers, making the model behave maliciously when specific inputs are given. Research on backdoor attacks explored various mechanisms to compromise LLMs. Hidden triggers were implanted into models using word-based, scene-based, and Retrieval-Augmented Generation (RAG) techniques, demonstrating how these methods could embed vulnerabilities into systems~\cite{jiao2024trustembodiedagentsexploring}. Additionally, visual-language models (VLMs) integrated into robotic systems were examined for potential exploits, revealing that adversaries could manipulate them to execute harmful actions in real-world environments~\cite{wang2025trojanrobotphysicalworldbackdoorattacks}.  In addition, jailbreak attacks in embodied LLM-based robots were achieved through voice-based user interactions, effectively bypassing safety and ethical constraints~\cite{zhang2024badrobotmanipulatingembodiedllms}. Another approach involved generating adversarial and meaningful simple word suffixes to influence embodied AI, enabling precise voice injections capable of causing harm in the physical world, affecting both environments and humans~\cite{lu2024poexpolicyexecutableembodied}. Also, both untargeted and targeted attacks were employed to execute jailbreaks in LLM-based embodied models, further highlighting their security vulnerabilities~\cite{liu2024exploringrobustnessdecisionleveladversarial}.

\section{Defense}\label{sec:4}

In this section, we explore various defense mechanisms for embodied navigation, which are generally classified into two main categories: physical defenses and model-based defenses. In particular, some defenses integrate elements from both categories. Table \ref{tab:defenses} provides an overview of representative defense types. 

\subsection{Physical Defense} 
Several prior studies explored defenses against patch attacks on pixels instead of specifically designing for embodied systems, employing both empirical strategies~\cite{Xu_2023_WACV,Wu_2024_CVPR} and certified approaches~\cite{xiang2021patchguardprovablyrobustdefense,xiang2024patchcureimprovingcertifiablerobustness}. One approach focused on ``detection and removal'', where adversarial purification was applied to mitigate the impact of adversarial patches~\cite{Xu_2023_WACV}. Another method introduced a detection framework targeting naturalistic adversarial patches with deceptive features~\cite{Wu_2024_CVPR}. Beyond empirical strategies, a small receptive field CNN was used to limit the number of features that adversarial patches could corrupt, thereby improving model robustness~\cite{xiang2021patchguardprovablyrobustdefense}. Subsequent work further enhanced both efficacy and robustness by refining these certified defenses~\cite{xiang2024patchcureimprovingcertifiablerobustness}.

Unlike previous passive defenses, active defense mechanisms were introduced for embodied navigation, utilizing recurrent feedback to actively counter adversarial patches~\cite{wu2024embodiedactivedefenseleveraging}. This approach leveraged environmental context, addressing misaligned adversarial patches in real-world 3D settings.

\subsection{Model-based Defense}

We name ``model-based defense'' to align with model-based attacks, as the defenses discussed here are specifically designed to counter the previously mentioned attacks. By maintaining this alignment, these defense strategies leverage model-driven mechanisms to effectively mitigate adversarial threats. The Embodied Active Defense (EAD) method was introduced to tackle adversarial patches in the 3D real world, actively integrating perception and action to interact with and adapt to the environment, thereby enhancing decision-making~\cite{wu2024embodiedactivedefenseleveraging}. To assess the safety of federated embodied agents, a real-time defense mechanism was developed by \cite{zhang2024navigationattackerswishbuilding}, implementing a Prompt-Based Aggregation (PBA) mechanism that detects malicious clients by analyzing vision-language alignment variance, thus providing more robust protection against federated learning attacks. Some studies focused on defense strategies for embodied navigation based on large language models. Various methods were evaluated to determine their effectiveness against backdoor attacks on embodied models~\cite{jiao2024trustembodiedagentsexploring}. Notably, directly deploying defense models (such as Llama-Guard-2, Llama-Guard-3, and Harmbench) has been a common approach. Both prompt-level and model-level defenses were explored to mitigate jailbreak attacks on embodied AI~\cite{lu2024poexpolicyexecutableembodied}.

\section{Evaluation}
In this section, we focus on the safety assessment issues in embodied navigation. Initially, we review safety-related datasets, followed by an organization of metrics used to evaluate safety. 

\subsection{Dataset}
Some of the benchmarks used in our work in Chapters \ref{sec:3} and \ref{sec:4} were not originally designed for embodied navigation (e.g. \cite{chen2024towards}). Additionally, some works have created their own datasets to meet their specific experimental needs \cite{yang2024hijackingvisionandlanguagenavigationagents}. In this section, we concentrate on the recent and representative datasets for the safety of embodied navigation.
\begin{table}[ht]
    \centering
    \resizebox{0.48\textwidth}{!}{%
    \begin{tabular}{@{} ccccc @{}}
        \toprule
        Defense & 
        \shortstack{Physical\\ Defense} &  
        \shortstack{Model-based\\ Defense} & 
        Attack Type & 
        Core Method \\
        \midrule
        \cite{Xu_2023_WACV}  
        & \checkmark & 
        & white-box
        & Detection and Remove \\
        \cite{Wu_2024_CVPR}  
        & \checkmark & 
        & black-box
        &  Feature Aligned Learning\\
         \cite{wu2024embodiedactivedefenseleveraging}  
        & \checkmark & \checkmark
        & white-box
        & Reinforcement learning \\
         \cite{zhang2024navigationattackerswishbuilding}  
        & & \checkmark
        & black-box
        & Federated learning \\
         \cite{lu2024poexpolicyexecutableembodied}  
        & & \checkmark
        & black-box
        & LLM Jailbreak \\
         \cite{jiao2024trustembodiedagentsexploring}  
        & & \checkmark
        & white-box
        & LLM Backdoor Attack \\
        \bottomrule
    \end{tabular}
    }
\caption{Different types of representative defenses. ``Attack Type'' refers to the category of attacks that this defense is designed to counter. }
\label{tab:defenses}
\end{table}
And we categorize the benchmarks based on the number of model parameters into those designed for classic models and those designed for LLMs.

\subsubsection{Datasets for classic models}
Based on the work of \cite{li2023benchmarkingassessingvisualnaturalness}, a physical attack naturalness dataset was constructed using human ratings and gaze data. Due to the limitations of virtual environments in replicating object interactivity and scene scale found in real-world settings, a photo-based 3D benchmark was later developed~\cite{kim2024realfredembodiedinstructionfollowing}. By integrating authentic scenes, objects, and room layouts, this benchmark allowed agents to better comprehend language instructions, complete household tasks, and operate in large-scale, multi-room real-world environments.
\begin{table*}[htbp]
\centering
\resizebox{\textwidth}{!}{%
\begin{tabular}{l|c| c c c}
\toprule
\textbf{Evaluation Dataset }& \textbf{Dataset Construction} & \textbf{\# Size} & \textbf{Evaluation Metric} &\textbf{ Domain
}\\
\midrule
\cite{li2023benchmarkingassessingvisualnaturalness} & Autonomous driving image & 2,688 images & Human-based &  Physical world attacks\\
\cite{kim2024realfredembodiedinstructionfollowing} & ALFRED\textsuperscript{1}\&Human annotation & 150 scenes & Formula-based &  Photo-realistic environments navigation\\
\cite{khanna2024goatbenchbenchmarkmultimodallifelong} & Real-world 3D scans from HM3DSem\textsuperscript{2} & 312 categories& Formula-based &  Multi-modal lifelong navigation\\
\cite{choi2024lotabenchbenchmarkinglanguageorientedtask} & ALFRED\&WAH\textsuperscript{3} & 308 tasks & Formula-based & Language oriented task planner\\
\cite{zhu2024earbenchevaluatingphysicalrisk} & GPT-4o generation& 2,636 samples & Formula-based &  Physical risk task planning\\
\cite{yin2024safeagentbenchbenchmarksafetask} & GPT-4 generation & 750 tasks & Model-based &  Safe task planning\\
\cite{wang2024divscenebenchmarkinglvlmsobject} &  GPT-4 generation\&Holodeck & 4,614 scenes & Formula-based &  LVLMS for object navigation\\
\cite{wang2025trojanrobotphysicalworldbackdoorattacks} &  GPT-4 generation\&Human annotation & 328 tasks & Formula-based &  MLLM navigation\\
\bottomrule
\end{tabular}%
}
\caption{Different types of representative evaluation datasets can be described by several aspects: ``Dataset Construction" details the process used to build the dataset; ``\# Size" indicates the size of the dataset; ``Evaluation Metric" represents the three different classification types; and ``Domain" denotes the scope within which the dataset is applied. }
\label{tab:dataset}
\end{table*}
Efforts to enhance embodied navigation benchmarks extended to diverse settings, including houses, gardens, restaurants, and offices. Objects within these environments were annotated with detailed physical and semantic attributes, with a focus on both reinforcement learning (RL) agents and safety concerns~\cite{li2024behavior1khumancenteredembodiedai}. Further advancements in multimodal lifelong navigation introduced a benchmark designed to challenge agents in open-vocabulary navigation tasks \cite{khanna2024goatbenchbenchmarkmultimodallifelong}. Agents were required to locate targets specified by category names, natural language descriptions, or images, contributing to the development of general-purpose navigation systems~.

\subsubsection{Datasets for LLMs}

With the development of LLMs, embodied LLM agents became more effective in interacting with people and making informed decisions in navigation. However, as previously discussed, LLMs remain vulnerable to jailbreak and backdoor attacks. In response, various benchmarks were introduced to evaluate LLM-based navigation systems, particularly focusing on recently developed datasets.
A benchmark suite was designed to automatically assess the task-planning capabilities of LLMs~\cite{choi2024lotabenchbenchmarkinglanguageorientedtask}. Each dataset sample provided natural language instructions and an environment to the planner. The simulator executed the planned actions and evaluated performance by comparing the final state with a predefined target condition. While its contributions, the dataset primarily focused on planning abilities rather than safety concerns.

To explore physical risks in embodied AI, dangerous scenarios were generated using LLMs and diffusion models, leading to an automated framework for risk assessment~\cite{zhu2024earbenchevaluatingphysicalrisk}. Various open-source and closed-source models were evaluated within this framework. However, safety analysis was mostly restricted to the input text, treating the embodied environment as an additional input rather than a core aspect of evaluation.
Recognizing this limitation, an alternative approach placed safety concerns at the center of evaluation, focusing on embodied agents that directly interact with the physical world rather than language models that only process text. ~\cite{yin2024safeagentbenchbenchmarksafetask} addressed ten common risks affecting humans and property, categorizing tasks into detailed tasks, abstract tasks, and long-horizon tasks to explore safety issues at various levels of abstraction and task duration.

Efforts to expand navigation-related datasets included the introduction of tasks requiring agents to navigate to various target objects across multiple scenarios. A dataset~\cite{wang2024divscenebenchmarkinglvlmsobject} covering 4,614 houses across 81 scenario types was constructed, utilizing Holodeck\textsuperscript{4} to generate textual descriptions of houses, while GPT-4 was employed to determine layout, style, and object placement. Textual annotations were later added to enrich the dataset’s contextual information.
In the realm of multimodal large-model embodied agents, an evaluation framework was established to assess capabilities across five different categories, including navigation~\cite{cheng2025embodiedevalevaluatemultimodalllms}. Task generation was powered by LLMs, with manual annotation and rigorous scene screening ensuring the dataset’s high quality and reliability. Notably, unlike traditional dataset creation methods, leveraging the generative power of LLMs in combination with human screening has emerged as a promising direction for developing high-quality embodied AI datasets.
\footnotetext[1]{\url{https://github.com/askforalfred/alfred}}
\footnotetext[2]{\url{https://aihabitat.org/datasets/hm3d-semantics/}}
\footnotetext[3]{\url{https://github.com/xavierpuigf/watch_and_help}}
\footnotetext[4]{\url{https://yueyang1996.github.io/holodeck/}}

\subsection{Metric}
The evaluation approaches for embodied navigation are diverse. Depending on the dataset and specific tasks, the methods can vary. Here, we primarily categorize the evaluation methods into three groups: human-based evaluation, formula-based evaluation, and model-based evaluation.
\subsubsection{Human-based evaluation}
Human-based evaluation is an assessment method that directly involves human judgment in evaluating a system’s performance. It is the simplest evaluation approach, ensuring both accuracy and reliability. \cite{li2023benchmarkingassessingvisualnaturalness} investigated the naturalness of physical-world attacks using human ratings and gaze data. Manual annotation was widely used in most studies to ensure the high quality of datasets, as demonstrated by \cite{kim2024realfredembodiedinstructionfollowing,cheng2025embodiedevalevaluatemultimodalllms,yin2024safeagentbenchbenchmarksafetask}. In some works, human evaluation was directly used to determine the correctness of the results (e.g., \cite{huang2022innermonologueembodiedreasoning}), with the \textit{success rate} serving as the primary evaluation metric.
\subsubsection{Formula-based evaluation}
Due to the costly and time-consuming nature of human-based evaluation, most benchmarks have begun shifting towards formula-based evaluation methods. This approach relies on predefined formulas and definitions to conduct the assessment. Here, we introduce some commonly used methods and formulas. In the benchmarks we mentioned, common metrics include: 
Success Rate (SR) \cite{yin2024safeagentbenchbenchmarksafetask,kim2024realfredembodiedinstructionfollowing}, 
Success weighted by Path Length (SPL) \cite{wang2024divscenebenchmarkinglvlmsobject,khanna2024goatbenchbenchmarkmultimodallifelong}, 
Success weighted by Episode Length (SEL) \cite{wang2024divscenebenchmarkinglvlmsobject}, 
and Goal-condition Success (GcS, abbreviated as GC) \cite{cheng2025embodiedevalevaluatemultimodalllms,kim2024realfredembodiedinstructionfollowing}.

An episode is considered successful if the target object appears in the agent's egocentric view and is within 1.5 meters of the agent. To maintain consistent notation, we denote the total number of episodes by \( M \) and index each episode by \( k \) (where \( k = 1, 2, \ldots, M \)). In this framework, \( s_k \) is a binary indicator of success (with \( s_k = 1 \) if the episode is successful, and \( s_k = 0 \) otherwise), \( d_k \) represents the length of the optimal (i.e., shortest) path to the target, and \( p_k \) is the length of the path traversed by the agent. For metrics based on episode length, \( d_k^a \) and \( p_k^a \) denote the number of actions along the optimal path and the agent's actual trajectory, respectively, while \( c_k \) indicates the number of goal conditions satisfied in episode \( k \), and \( C \) is the total number of predefined goal conditions. Using these definitions, the metrics are computed as follows: the SR is given by \( SR = \frac{1}{M} \sum_{k=1}^{M} s_k \); the SPL is calculated as \( SPL = \frac{1}{M} \sum_{k=1}^{M} s_k \cdot \frac{d_k}{\max(d_k,\, p_k)} \); the SEL is determined by \( SEL = \frac{1}{M} \sum_{k=1}^{M} s_k \cdot \frac{d_k^a}{\max(d_k^a,\, p_k^a)} \); and the GC is computed as \( GC = \frac{1}{M} \sum_{k=1}^{M} \frac{c_k}{C} \). If a system is considered safe, the SR, SPL, SEL, and GC metrics should be as high as possible, reflecting its ability to perform tasks efficiently, securely, and reliably. When considering navigation efficiency simultaneously, time \( t \) is also an important evaluation metric.

\subsubsection{Model-based evaluation}
Beyond leveraging large models for data generation, some benchmarks also adopt model-based evaluation for assessing performance. In particular, abstract tasks often allow for multiple valid execution strategies rather than a single definitive solution. To address this variability, \cite{yin2024safeagentbenchbenchmarksafetask} employed GPT-4 to evaluate the plausibility and effectiveness of execution plans generated by the model. This approach ensures that the proposed plans align with task objectives and maintain coherence. Building on these evaluations, performance was further quantified by computing both the success rate and the probability of rejection, providing a systematic way to assess the model’s robustness and safety.

\section{Future Directions}\label{sec:6}
In this section, we discuss several unresolved challenges in the safety of embodied navigation, highlighting key issues that remain to be addressed. We aim to offer valuable insights and propose potential research directions that could foster the development of safer and more efficient embodied navigation systems in the future.
\subsection{Potential Attack Methods}

There are three potential research directions for adversarial attacks:
\textbf{(a) Enhancing Robustness in Dynamic Environments}:  
Some existing studies primarily focus on developing attack and defense strategies for specific agent tasks, demonstrating the effectiveness of their proof-of-concept approaches. However, these methods often face significant limitations when applied to real-world, complex environments. For instance, object-trigger-based attacks require precise visual consistency across multiple viewpoints, making them less effective in dynamic settings with varying perspectives. \textbf{(b) Expanding Attack Types}: Certain attack strategies operate under a black-box assumption, where the attacker lacks prior knowledge of the model’s internal mechanisms, thereby limiting their applicability in scenarios that require more fine-grained adversarial optimization. As a result, expanding the scope of attack methodologies, including white-box attacks, remains a critical avenue for further research.  \textbf{(c) Adversarial Attacks on Multimodal Models}: Existing model-based attack research has primarily focused on large language models. However, with the rapid advancement of multimodal large models, traditional attack paradigms may not seamlessly transfer to multimodal settings. Therefore, investigating attack strategies specifically designed for multimodal models, such as cross-modal perturbations, represents an important yet unsolved direction. 

\subsection{Robust Defense Strategies}
Currently, research on physical defenses (e.g., patch-based defenses) in embodied navigation remains limited, leaving significant room for further exploration. Existing defense mechanisms are often designed for other tasks and may not fully address the unique challenges of embodied navigation, such as real-time decision-making and continuous interaction with dynamic environments. In LLM-based navigation systems, models may inherit vulnerabilities from text-based models, including sensitivity to prompts and susceptibility to adversarial text. Strengthening the defense capabilities of LLMs in embodied navigation, such as developing more robust language understanding mechanisms, integrating adversarial-resistant knowledge injection, or enforcing multimodal consistency constraints, remains a critical research direction. Additionally, due to the real-time interactive nature of embodied systems, runtime monitoring techniques could be explored as a defense strategy to dynamically identify and counteract adversarial threats during execution.
\subsection{Reliable Evaluation}
Current research often focuses primarily on qualitative assessments without conducting rigorous quantitative experiments. Even studies that incorporate quantitative evaluations typically rely on different datasets or purpose-built datasets for experimentation, lacking direct comparisons with other security-related approaches. As a result, a more systematic analysis of security remains an important research direction. Moreover, evaluation methods can be transitioned to model-based approaches, such as leveraging GPT-4 or other large models to assess accuracy, which can automate the process and reduce human effort. Additionally, different AI tasks, such as visual exploration and LLM-based question answering, adopt distinct evaluation metrics, making direct comparisons between studies challenging. Therefore, future research should focus on developing a more unified evaluation framework to ensure fairness and interpretability across different tasks. Furthermore, exploring new evaluation methods, such as integrating multiple metrics or incorporating human feedback, could further enhance the reliability and quality of AI task evaluation.
\subsection{Verification Techniques }
Current research lacks a systematic approach to the verification of embodied navigation, making this a promising direction for further exploration. One potential avenue is quantifying the range of input perturbations that do not affect the model’s output, thereby establishing robustness thresholds. Additionally, computing theoretical bounds under different conditions presents another valuable research direction, as these bounds not only provide guidance for safety research but also serve as key metrics for evaluating system robustness. These efforts would contribute to enhancing the reliability and security of embodied navigation while providing theoretical support for the development of more robust navigation algorithms.
\section{Conclusion}
In this paper, we present a detailed overview of the safety of embodied navigation. We review recent research advancements from three key perspectives: attack strategies, defense mechanisms, and evaluation methodologies. Finally, based on the current state of research, we identify several promising directions for future investigation, aiming to foster the development of safer and more robust embodied navigation systems.

\bibliographystyle{named}
\bibliography{ijcai25}

\end{document}